\documentclass[journal]{IEEEtran}
\ifCLASSINFOpdf
\else
\fi
\usepackage[pdftex]{graphicx}
\usepackage{multirow}
\usepackage{amsmath}

\begin{document}
%
\title{Deep Feature Consistent Deep Image Transformations: Downscaling, Decolorization and HDR Tone Mapping}
%
%
%

%
\author{Xianxu~Hou,
        Jiang~Duan,
        and~Guoping~Qiu
\thanks{X. Hou is with the School of Computer Science, University of Nottingham Ningbo China and a visiting student in the College of Information Engineering, Shenzhen University, China. email: xianxu.hou@nottingham.edu.cn}

\thanks{J. Duan is with the School of Economic Information Engineering, Southwestern University of Finance and Economics, China. email: duanj\_t@swufe.edu.cn}

\thanks{G. Qiu is with the College of Information Engineering, Shenzhen University, China and with the School of Computer Science, the University of Nottingham, United Kingdom. email: qiu@szu.edu.cn and guoping.qiu@nottingham.ac.uk}}
\maketitle

\begin{abstract}
Building on crucial insights into the determining factors of the visual integrity of an image and the property of deep convolutional neural network (CNN), we have developed the Deep Feature Consistent Deep Image Transformation (DFC-DIT) framework which unifies challenging one-to-many mapping image processing problems such as image downscaling, decolorization (colour to grayscale conversion) and high dynamic range (HDR) image tone mapping. DFC-DIT uses one CNN as a non-linear mapper to transform an input image to an output image following the deep feature consistency principle which is enforced through another pretrained and fixed deep CNN. Unlike applications in related  literature, none of these problems has a known ground truth or target for a supervised learning system. For each problem, we reason about a suitable learning objective function and develop an effective solution under the DFC-DIT framework. This is the first work that uses deep learning to solve and unify these three common image processing tasks. We present experimental results to demonstrate the effectiveness of the DFC-DIT technique and its state of the art performances.

\end{abstract}


%
\IEEEpeerreviewmaketitle

\begin{figure*}
  \includegraphics[width=\textwidth]{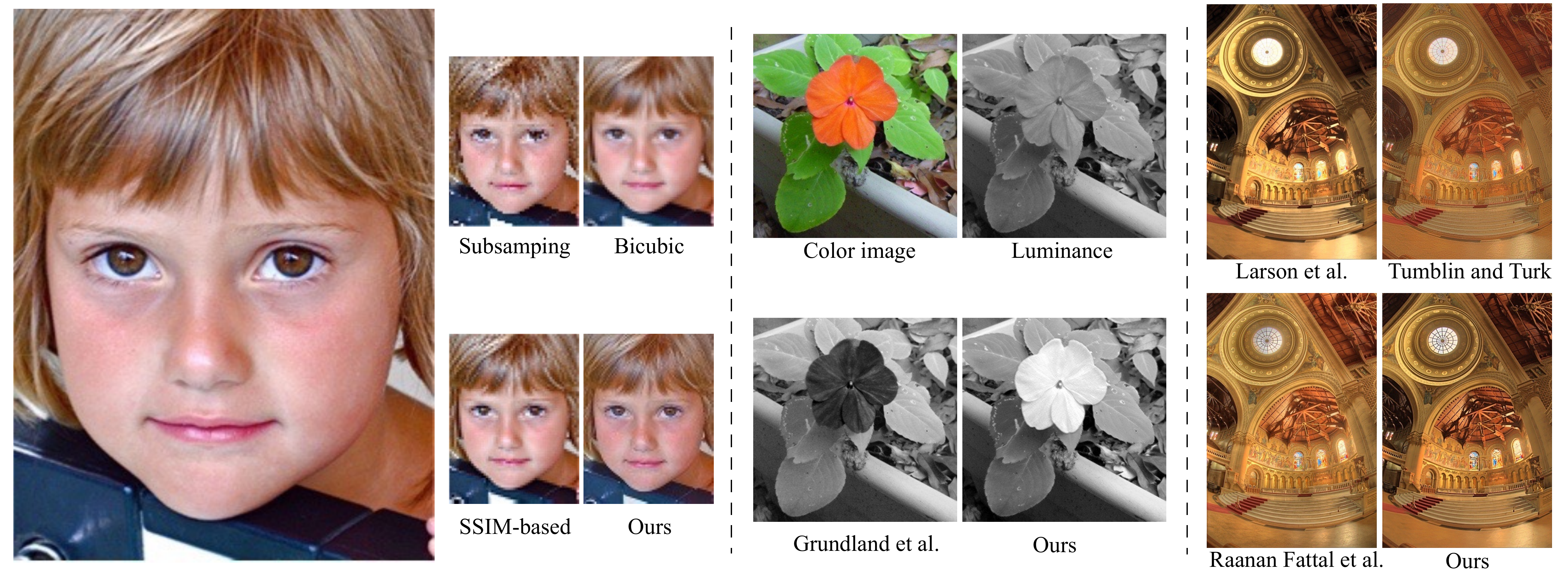}
  \caption{Examples of classic image transformation tasks. Image downscaling (left) where we show results of our method, two traditional methods (subsampling and bicubic) and a state-of-the-art SSIM-based method \cite{oeztireli2015perceptually}. Decolorization (middle) where we show results of our method, the Luminance channel and a state-of-the-art method \cite{grundland2007decolorize}. HDR image tone mapping (right) where we show results of our method and 3 methods from the literature \cite{larson1997visibility, tumblin1999lcis,fattal2002gradient}.}
  \label{fig:teaser}
\end{figure*}

\section{Introduction}
Many classic image processing tasks (Fig. \ref{fig:teaser}) can be framed as image transformation, where an input image is transformed to an output image based on a given criterion. In this paper, we consider three image transformation tasks: image downscaling, image decolorization (color to grayscale conversion), and high dynamic range (HDR) image tone mapping. Downscaling image operations are widely used today to allow users to view a reduced resolution image that preserves perceptually important details of its original megapixel version. Decolorization aims to convert a color image to a grayscale image which will preserve the visual contrasts of the original colour version. Another image processing task is HDR image tone mapping. HDR images contain a much higher bit depth than standard image formats and can represent a dynamic range closer to that of human vision. The goal of HDR tone mapping is trying to faithfully reproduce the appearance of the high dynamic range image in display devices with limited displayable range. 

These three seemingly disparate image processing tasks have a similar objective of outputting a reduced information version which will maximally convey important perceptual details of the original image. All these tasks face similar technical challenges. The transformations are one-to-many mapping and there are no known targets, an image can be transformed to an arbitrary number of plausible outputs. The transformation criterion, e.g., to preserve the perceptual details and contrasts of the original image, etc., are qualitative and subjective. There is no well-defined mathematical objective function to describe the transformation criterion and this makes it difficult to find a canonical computational solution to these problems.

In this paper, we take advantage of recent developments in deep convolutional neural networks (CNNs) and have developed a deep feature consistent deep image transformation (DFC-DIT) framework in which we train a deep CNN to transform an input image to an output image by keeping the deep features of the input and output consistent through another pre-trained (and fixed) deep CNN. We show that common traditional image processing tasks such as downscaling, decolorization and HDR tone mapping can be unified under the DFC-DIT framework and produce state-of-the-art results. To the best knowledge of the authors, this is the first work that successfully uses deep learning to solve downscaling, decolorization and HDR tone mapping problems in a unified framework.

\begin{figure*}
  \includegraphics[width=\textwidth]{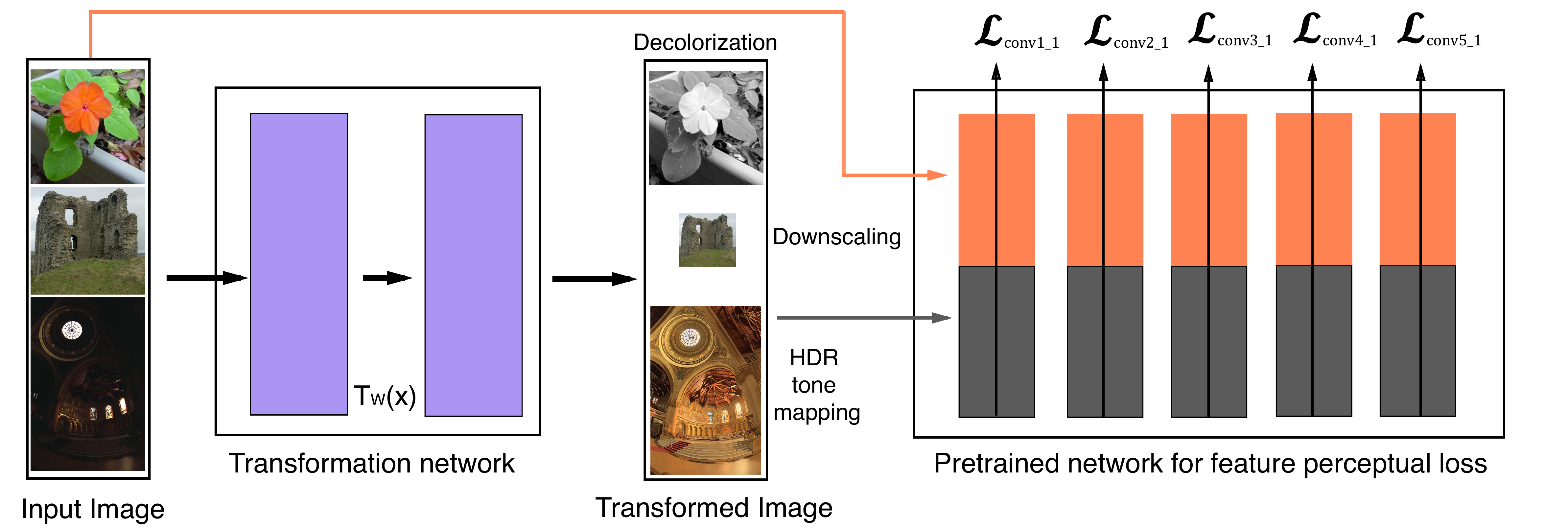}
  \caption{The Deep Feature Consistent Deep Image Transformation (DFC-DIT) framework. A convolutional neural network transforms an input to an output. A pretrained deep CNN is used to compute feature perceptual loss for the training of the transformation network.}
  \label{fig:overview}
\end{figure*}

The new DFC-DIT framework is built on two crucial insights, one into the visual appearance of an image and the other into the properties of the deep convolutional neural networks. From image quality measurement literature, it is known that the change of spatial correlation is a major factor affecting the visual integrity of an image \cite{995823}. Research in deep learning has shown that the hidden layers of a convolutional neural network can capture a variety of spatial correlation properties of the input image \cite{hou2016deep}. As one of the most important objectives of many image processing tasks such as the three studied in this paper is to maximally preserve the visual integrity of the input, it is therefore crucial to keep the spatial correlations of the output consistent with those of the input. As the deep features (i.e., hidden layers' outputs) of a CNN capture the spatial correlations of the input image, we can therefore employ a (pre-trained and fixed) CNN and use its deep features to measure the spatial correlations of an image. Therefore, the goal of preserving the visual integrity of the input is equivalent to keeping the spatial correlations of the output consistent with that of the input, which in turn is equivalent to keeping their deep features consistent. Based on these key insights, we have successfully developed the DFC-DIT image processing framework (see Fig. \ref{fig:overview}).

The rest of the paper is organized as follows. We briefly review related literature in section 2. Section 3 presents the DFC-DIT framework and its application to three important image processing tasks, i.e., spatial downscaling, decolorization and high dynamic range image tone mapping. Section 4 presents experimental results which show that DFC-DIT stands out as a state-of-the-art technique. Finally we present a discussion to conclude the paper.

\section{Related Work}
\textbf{Image downscaling.} Classical image downscaling techniques usually involve processing the input images by applying a spatially constant low-pass filter, subsampling, and reconstructing the result to prevent aliasing in the reconstructed signal. Approximations to the theoretically optimum sinc filter such as the Lanczos filter, and other filters (e.g., bilinear and bicubic) have been developed and used in practice. However the filtering kernels of these methods do not adapt to the image content. A recent content-adaptive technique \cite{kopf2013content} is proposed to overcome the above shortcoming by adapting the shape and location of every kernel to the local image content and demonstrates better downscaling results. A better depiction of the input image was proposed \cite{oeztireli2015perceptually} by formulating image downscaling as an optimization problem with the structural similarity (SSIM) \cite{wang2004image} as the perceptual image quality metric. In addition convolutional filters \cite{Weber2016} are used to preserve visually important details in downscaled images.

\textbf{Decolorization.}
Decolorization aims to convert color images into grayscale images while preserving structures and contrasts as much as possible. The baseline method is to extract the luminance channel of a given color image from the RGB channels. However it could fail to express salient structures of the color image because of the fixed weights to combine RGB channels. Other more advanced techniques are proposed to obtain better results by either focusing on local contrasts or global contrasts. Local contrasts \cite{gooch2005color2gray,smith2008apparent} use different mapping functions in different local regions of the image, while global contrasts \cite{rasche2005re,ancuti2011enhancing,lu2012contrast,qiu2008contrast} are designed to produce one mapping function for the whole image. \cite{song2013decolorization} takes into account multi-scale contrast preservation in both spatial and range domain and uses bilateral filtering to mimic human contrast perception. \cite{lu2014contrast} used a bimodal objective function to alleviate the restrictive order constraint for color mapping. In addition image fusion based strategy \cite{ancuti2010image} is proposed for image and video decolorization.

\textbf{HDR image tone mapping.}
HDR image tone mapping aims to reproduce high dynamic range radiance maps in low dynamic range reproduction devices. 
Tone mapping operators can be classified as global operators and local operators. Global operators \cite{tumblin1993tone,ward1994contrast,qiu2007learning} usually employ the same mapping function for all pixels and can preserve the intensity orders of the original scenes to avoid ``halo'' artifacts, however the global operators will generally cause loss of details in the mapped image. In contrast, local operators \cite{tumblin1999lcis,durand2002fast,reinhard2002photographic} use mapping functions which vary spatially across the image. 
Most local operators employ a pipeline to decompose an image into different layers or scales and then recompose the mapped results from various scales after contrast reduction. However, the major shortcoming of local operators is the presence of haloing artifacts. In addition, global operator is used in the local regions to reproduce local contrast and ensure better quality \cite{duan2010tone}. What's more, an up-to-date, detailed guide on the theory and practice of high dynamic range imaging is included in the book \cite{Banterle2011}, which also provide MATLAB code for common tone mapping operators. In this paper, we use their code to reproduce previous methods.

\textbf{Image quality metrics.}
The choice of image quality metric is essential for image transformation tasks. Standard pixel-by-pixel measurement like mean square error is problematic and the resultant images are often of low quality. This is because the measurement is poorly correlated with human perception and can not capture the perceptual difference and spatial correlation between two images. Better metrics have been proposed for image quality assessment in recent years. Structural similarity (SSIM) index \cite{wang2004image} is one of the most popular metrics, which computes a matching score between two images by local luminance, contrast, and structure comparisons. It has been successfully used for image downscaling \cite{oeztireli2015perceptually}, image denoising \cite{wang2004image} and super-resolution \cite{zhou2014single}. 

\textbf{Relevant deep learning/CNN literature.}
Recently, there has been an explosion of publications on deep learning/CNN, we here briefly review the most closely related publications to our current work. A number of papers have successfully generated high-quality images based on the high-level features extracted from pretrained deep convolutional neural networks. By optimizing individual deep features \cite{yosinski2015understanding,szegedy2013intriguing,nguyen2015deep,hou2016deep}, better visual quality images can be generated, which in turn can help understand the learned representations of deep networks. Additionally \cite{gatys2015neural} have achieved style transfer by minimizing content and style reconstruction loss which are also based on features extracted from deep networks. Other works try to train a feed-forward network for real-time style transfer and super-resolution \cite{johnson2016perceptual}. Different loss functions are compared for image restoration with neural networks \cite{7797130}. In addition image-to-image translation framework \cite{isola2016image} are proposed to generate high quality images based on adversarial training.

It is worth noting that the downscaling problem studied in this paper has the opposite goal to super-resolution. Deep CNN based super-resolution training data has a unique corresponding target for a given input image and is a many-to-one mapping. The downscaling operation, however, is a one-to-many mapping; for a given input, there is no known target in the training data. Therefore, existing end to end super-resolution learning \cite{dong2014learning,kim2016accurate,johnson2016perceptual} and other similar CNN based image processing techniques such as colorization  \cite{zhang2016colorful,larsson2016learning} cannot be directly applied to the problems studied in this paper.

\begin{figure*}
  \includegraphics[width=\textwidth]{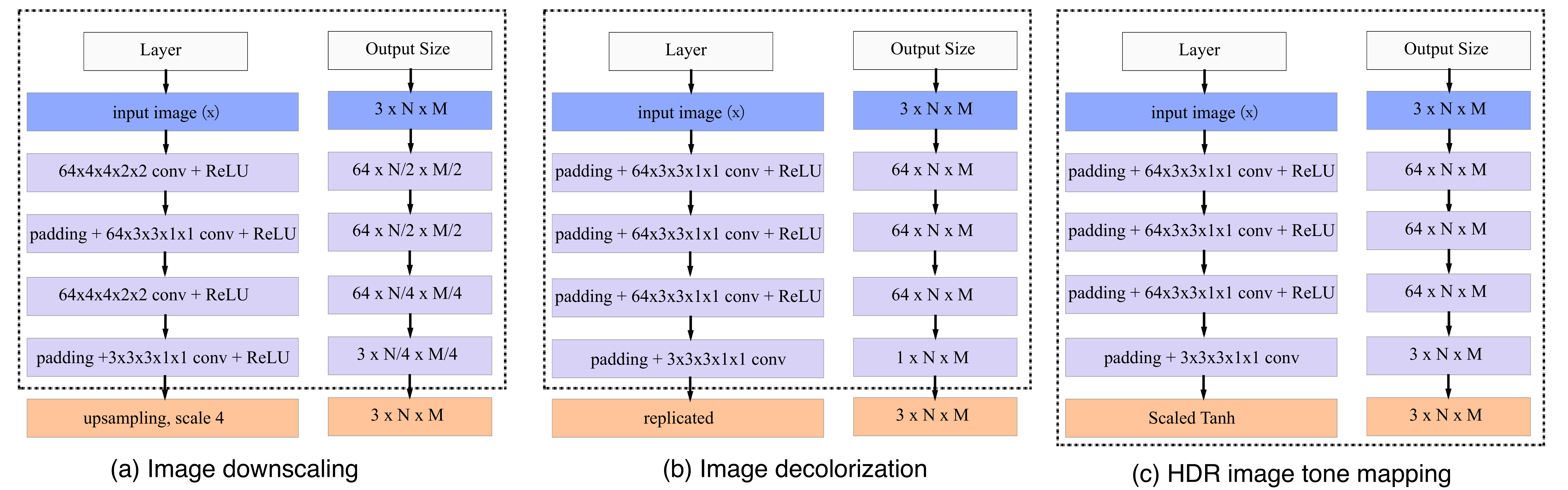}
  \caption{Transformation neural network architecture for image downscaling, decolorization and HDR image tone mapping.}
  \label{fig:architecture}
\end{figure*}

\section{Method}
We seek to train a convolutional neural network as a non-linear mapper to transform an input image to an output image following what we call the deep feature consistent principle. 
The schematic is illustrated in Fig. \ref{fig:overview}. Our system consists of two components: a transformation network $T_W(x)$ and a loss network $\Phi(x)$. The transformation network is a convolutional neural network parameterized by weights $W$, which transforms an input image $x$ to an output image $\hat{x}$, i.e. $\hat{x} = T_W(x)$. The other component is the loss network $\Phi$ which is a pretrained deep convolutional neural network to help define the feature perceptual loss function for training $T_W(x)$. We feed both the original image $x$ and the transformed image $\hat{x}$ to $\Phi$ and compute the feature perceptual loss $\mathcal{L}(x, \hat{x})$. 
Training $T_W(x)$ is to find the weights $W$ that minimize $\mathcal{L}(x, \hat{x})$, i.e.

\begin{equation}
W^* = \arg \min_{W} E_x [\mathcal{L}(x, T_W(x))]
\end{equation}

Equation (1) can be seen as an extension of the concept of perceptual loss, e.g. \cite{johnson2016perceptual} and others. However, the three new applications we consider here are very different from those studied by others. These extensions are non-trivial and non-obvious; each requires in-depth understanding of the problem and ingenuity that cannot be readily derived from existing works. Unlike previous applications, none of our problems has a known ground truth or target for a supervised learning network. Instead, we have to reason about the suitable target and develop solutions to construct the perceptual loss for each application accordingly. In downscaling, we created a perceptual loss to match two images with different shapes (sizes). In colour2gray, we constructed a perceptual loss to match two images with different number of colour channels. In HDR tone mapping, we introduced a perceptual loss to match two images with different dynamic ranges. 

\subsection{Deep Feature based Feature Perceptual Loss}

As alluded to earlier, the spatial correlation of an image is a major determining factor of the visual integrity of an image. The goal of image transformation in Fig. \ref{fig:overview} and the tasks in Fig. \ref{fig:teaser} is to ensure $\hat{x}$ preserves the visual integrity of $x$. This can be alternatively stated as making the spatial correlations in $\hat{x}$ consistent with those in $x$. Instead of using handcrafted functions to describe an image's spatial correlations, we make use of a pretrained deep CNN. The hidden layers outputs, which we call deep features, capture the spatial correlations of the input image. 

Specifically, let $\Phi_i(x)$ represent the $i^{th}$ hidden activations when feeding the image $x$ to $\Phi$. If the $i^{th}$ is a convolutional or ReLU layer, $\Phi_i(x)$ is a feature map of shape [$C_i$, $W_i$, $H_i$], where $C_i$ is the number of filter for the $i^{th}$ convolutional layer, $H_i$ and $W_i$ are the height and width of the given feature map respectively. The feature perceptual loss $\mathcal{L}_i(x, \hat{x})$ for a given layer of two images $x$ and $\hat{x} = T_W(x)$ is defined as the normalized Euclidean distance between the corresponding 3D feature maps. The final loss $\mathcal{L}_i(x, T_W(x))$ is the total loss of different layers. 

\begin{equation}
\footnotesize
\mathcal{L}_i(x,  T_W(x)) =  \frac{1}{C_iW_iH_i}  \sum_{c=1}^{C_i}  \sum_{w=1}^{W_i}  \sum_{h=1}^{H_i} (\Phi_i(x)_{c,w,h} - \Phi_i( T_W(x))_{c,w,h})^2
\end{equation}

\begin{equation}
\footnotesize
\mathcal{L}(x,  T_W(x)) = \sum_{i} \mathcal{L}_i(x,  T_W(x))
\end{equation}

It is worth noting that $\Phi$ is pre-trained and fixed during the training of $T_W(x)$, it is used as convolutional filters to capture the spatial correlations of the images.

\subsection{Transformation Networks Architecture}
The transformation networks are convolutional neural networks based on the architecture guidelines from VGGNet \cite{simonyan2014very} and DCGAN \cite{radford2015unsupervised}, and the details of the architecture vary with different image transformation tasks (Fig. \ref{fig:architecture}).

\begin{figure}
\includegraphics[width=8cm]{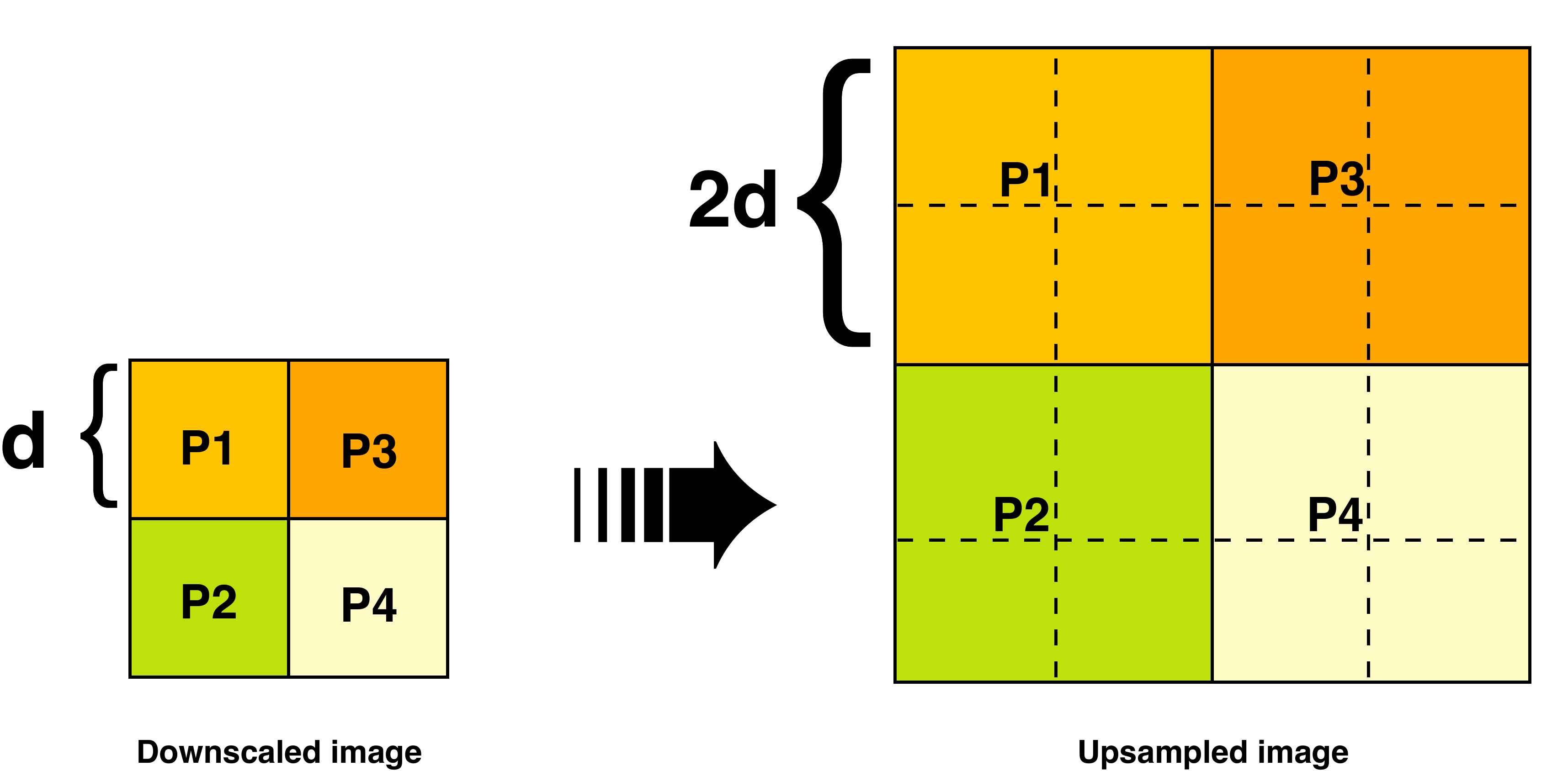}
  \caption{Nearest neighbor upsampling for the transformed image. The upsampled image contains the same amount of information as the downscaled image and is the same size as the original input image.}
  \label{fig:upsampling}
\end{figure}

\textbf{Image downscaling.}
For image downscaling we use strided convolutions to construct the networks with 4 x 4 kernels. The stride is fixed to be 2 x 2 to achieve in-network downsampling instead of deterministic spatial functions such as max pooling and average pooling. The ReLU layer is used after the first convolutional layer as non-linear activation function. Thus after two strided convolutions, the size of the input image can be downscaled to 1/4. In order to compute the feature perceptual loss we need to make sure that the transformed image and the original image have the same shape. In our experiments we apply a 2D upsampling of a factor of 4 over every channel of the transformed output (see Fig. \ref{fig:upsampling}), thus upscaling the downscaled image back to the same size as the original input. The nearest neighbor upsampler is chosen to ensure the upsampled image has the same information as the downscaled image. Thus we can feed the upscaled version and the original image into the loss network to compute the feature perceptual loss.

\textbf{Image decolorization.}
The image decolorization transformation only affects the color of the input images, and there is no need to incorporate downsampling architecture in the network. We use 3 x 3 kernels with 1 x 1 stride for all the convolutions. In addition, each feature map of a given input is padded by 1 pixel with the replication of the input boundary before the convolution operation. Thus the convolutional layers do not change the size of the input image. Like the image downscaling network we use ReLU layer after the first convolutional layer, but only a single filter for the last convolution to represent the transformed grayscale image. What we desired is the deep feature consistency of the decolorized output and the original image. 
We replicate the single channel of the decolorized output to a 3 channel color image (3 channels are identical), which is then fed to the loss network $\Phi(x)$ to calculate the feature perceptual loss with the original input. This is designed to ensure the replicated 3 channel color image have the same amount of information as the decolorized output.

\textbf{HDR image tone mapping.}
The network architecture for HDR image tone mapping is similar to the one used in image decolorization above. We use replication to pad the input boundary, and all the convolutions are 3 x 3 kernels with 1 x 1 stride. The difference is that 3 filters are needed for the last convolutional layer for reproducing a color image. The output layer is a scaled Tanh layer, restricting the pixel value of the transformed image to the displayable range [0, 255] from a high dynamic range. During the training we seek the deep feature consistency of the tone mapped and the original high dynamic range image. Specific implementation details of each of the applications are described in the experiments section.

\section{Experiments}
We present experimental results on three image transformation tasks: image downscaling, image decolorization and HDR image tone mapping to demonstrate the effectiveness of our method. We also investigate how the feature perceptual loss constructed with different hidden layers of the loss network affects the performances. 

\subsection{Training Details}
Our image downscaling and decolorization transformation CNNs are trained offline using Microsoft COCO dataset released in 2014 \cite{lin2014microsoft}, which is a large-scale dataset containing 82,783 training images. We resize all the image to 256 $\times$ 256 as the final training data, and train our models with a batch size of 16 for 10 epochs over all the training images. Once the transformation CNN is trained, it can be used to perform downscaling or decolorization.

For HDR image tone mapping, the transformation CNN is trained online, i.e., an HDR image is compressed using the transformation CNN trained with its own data. The practical consideration is that it is difficult to collect large enough training dataset. With large enough collection of training data, the model can also be trained offline. 

For training, Adam \cite{kingma2014adam} method is used for stochastic optimization with a learning rate of 0.0002. A pretrained 19-layer VGGNet \cite{simonyan2014very} is used as loss CNN $\Phi$ to compute feature perceptual loss which is fixed during the training of the transformation CNN. When constructing the feature perceptual loss for a pretrained network, the first step is to decide which layer (layers) should be used. Unlike image generation works \cite{johnson2016perceptual,hou2016deep} using ReLU layers, we use convolution layers for feature extraction. This is because the ReLU activation is just the corresponding convolutions output thresholded at 0, the convolutions could contain more subtle information when compared with ReLU output. Specifically we experiment feature perceptual loss by using convolutional layer conv1\_1, conv2\_1, conv3\_1, conv4\_1 and conv5\_1 for comparison. Our implementation is built on open source machine learning framework Torch \cite{collobert2011torch7}.

\subsection{Image Downscaling.}
Image downscaling is trying to transform a high-resolution input image to a low-resolution output image. In our experiments we focus on the $\times$ 1/4 image downscaling similar to previous works \cite{oeztireli2015perceptually,kopf2013content}. This seemingly simple routine image operation is actually a technically challenging task because it is very difficult to define the correct low-resolution image. As already discussed, this is a one-to-many mapping and there are many plausible solutions. Based on our DFC-DIT framework, we ensure that the downsampled image and the original image will have similar deep features which means that the output will maintain the spatial correlations of the original image thus keeping the visual integrity of the original image. 

\begin{figure*}
  \includegraphics[width=\textwidth]{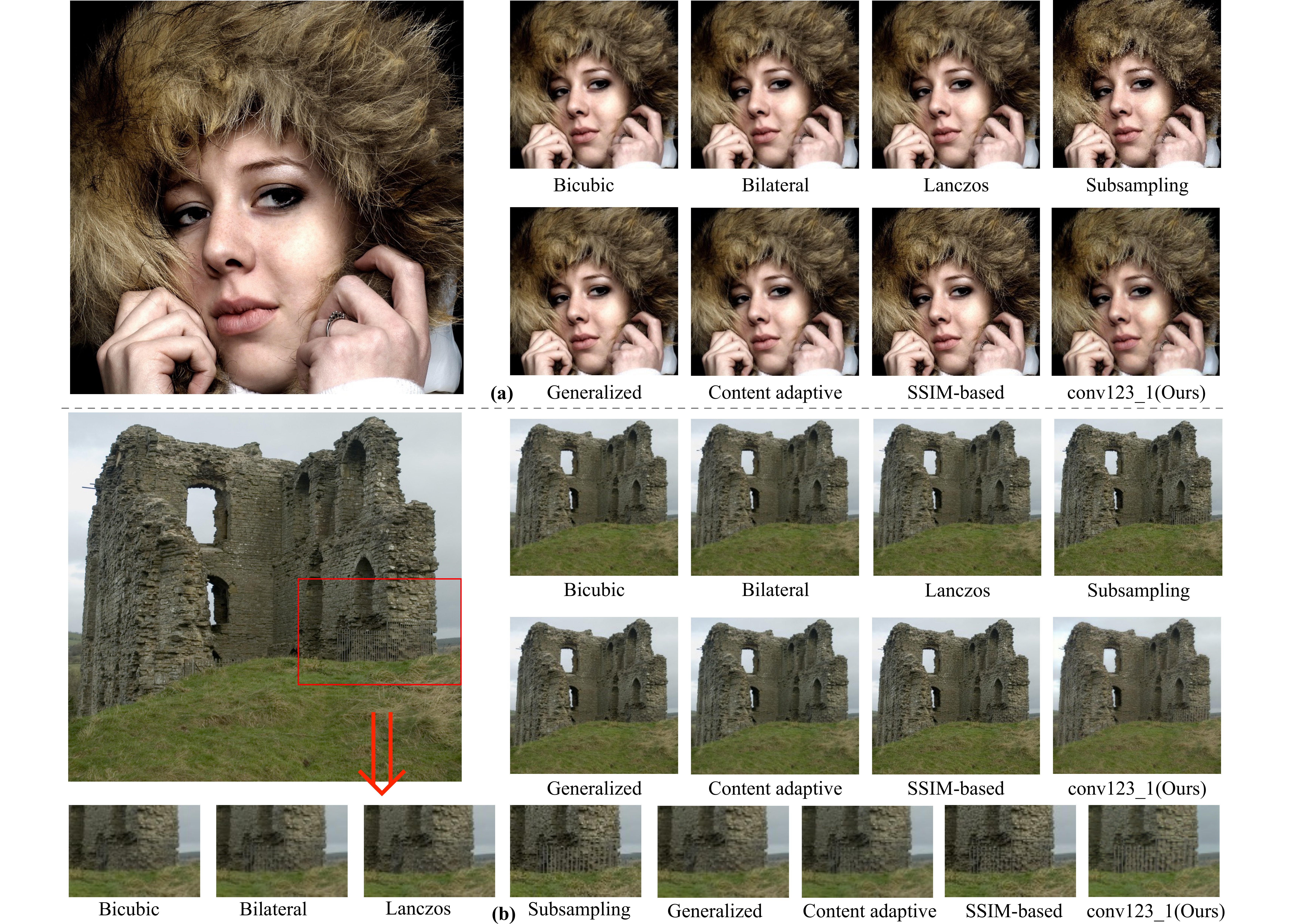}
  \caption{A comparison of natural images downscaled by different methods. The results are downscaled by a factor of $\times$ 1/4 while the original inputs are resized for better display. For each image, results of common filters such as Bicubic, Bilateral, Lanczos and Subsampling are shown in the first row. Results of recent methods, generalized sampling \cite{nehab2011generalized}, content-adaptive\cite{kopf2013content} and SSIM-based downsampling \cite{oeztireli2015perceptually} and ours are shown in the second row. Our conv123\_1 results are produced by a model trained with a combined loss of convolutional layers conv1\_1, conv2\_1, and conv3\_1. The bottom row of the second image shows a local region of the downscaled image by different methods. All the images are courtesy of \cite{oeztireli2015perceptually}. The results are best viewed in native resolution electronically.}
  \label{fig:downscaling_other}
\end{figure*}

\begin{figure*}
\includegraphics[width=\textwidth]{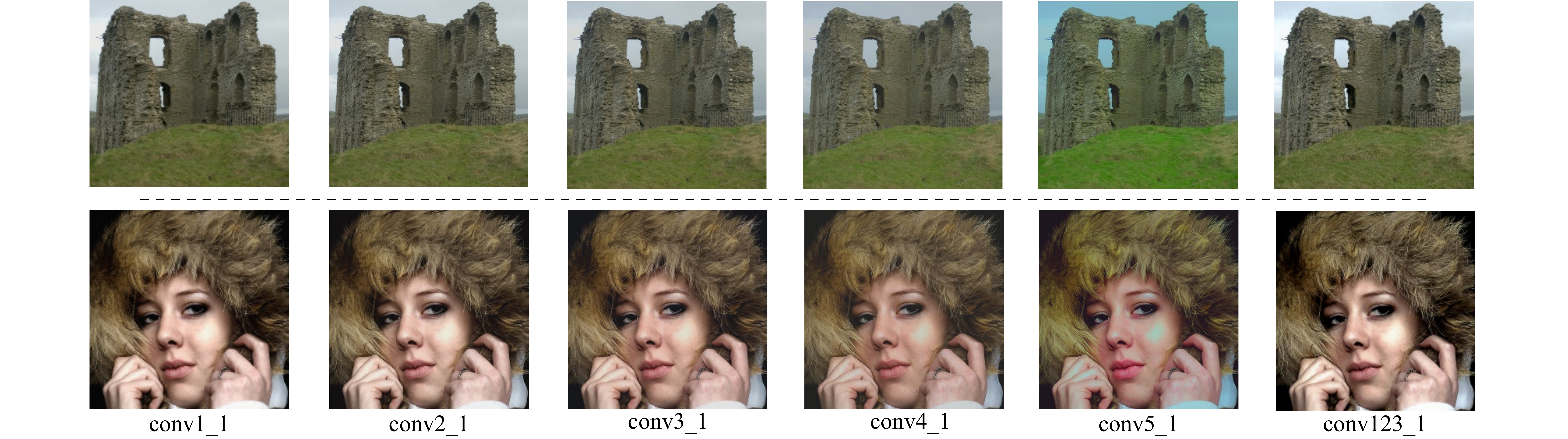}
  \caption{A comparison of natural images downscaled with the DFC-DIT framework with different levels of feature perceptual loss. The examples, from left to right, are $\times$ 1/4 downscaling results with perceptual losses computed with individual hidden layers of VGGNet (from layer 1 to layer 5). The last column is the results based on a perceptual loss combining the first 3 layers.}
  \label{fig:downscaling}
\end{figure*}

\textbf{Qualitative results}
Although our network is trained on images of shape 256 $\times$ 256, it can be adapted to any image sizes because of its fully convolutional architecture. After training, we evaluate our method on the testing images from \cite{oeztireli2015perceptually}. We first show the qualitative examples and compare our results with other state-of-the-art methods. We then evaluate how perceptual losses constructed at different convolutional layers affect the performances.

Fig. \ref{fig:downscaling_other} shows qualitative examples of our results, other common techniques and state-of-the-art methods. We only show results of downscaling by a factor of $\times$ 1/4, the original images are resized for better display. We can see that bicubic filter is known to lead to oversmoothing results and cannot preserve fine structures such as the fence area highlighted by the red rectangle (Fig. \ref{fig:downscaling_other}(b)). 
Other filters such as bilateral filter and Lanczos filter achieve sharper downscaled results, however these filters are also problematic. Bilateral filter can lead to ringing artifacts (the hair in Fig. \ref{fig:downscaling_other}(a)), and Lanczos filter could not preserve small-scale features such as the fence area in Fig. \ref{fig:downscaling_other}(b). More advanced methods such as generalized sampling \cite{nehab2011generalized}, and content-adaptive downscaling \cite{kopf2013content} and SSIM-based downscaling \cite{oeztireli2015perceptually} could produce better results, but still cannot preserve all perceptually important details. In contrast our method 
trained by a feature perceptual loss constructed using layer conv1\_1, conv2\_1 and conv3\_1 deep features can capture important fine textures and produce better transformed results, visually closer to the original high-resolution inputs. From Fig. \ref{fig:downscaling_other}(b), the fine textures of the fence area can be seen clearly in the downscaled image. Although simple (nearest neighbor) subsampling can also achieve sharper images, the results are sometimes noisy and suffer from aliasing (see the hair in Fig. \ref{fig:downscaling_other}(a)). Our algorithm avoids both oversmoothing and aliasing problems and produces a crisp and noise-free image. These results demonstrate that by keeping the deep features of the downscaled image consistent with those of the original can indeed preserve the visual integrity of the input.

\textbf{Deep feature consistency at different layers.} Fig. \ref{fig:downscaling} shows results of DFC-DIT downscaled images using perceptual losses computed using conv1\_1, conv2\_1, conv3\_1, conv4\_1 and conv5\_1 layer of the VGGNet individually. We find that keeping the deep feature consistent at these individual layers can in general preserve the original texture or content well. However for the high level layers, the downscaled images could lose detailed pixel information such as pixel color. For example, results of conv4\_1 and conv5\_1 in Fig. \ref{fig:downscaling} have higher color contrasts. We also found that by combining the first three layer deep features in general works very well.

\begin{figure*}
\includegraphics[width=\textwidth]{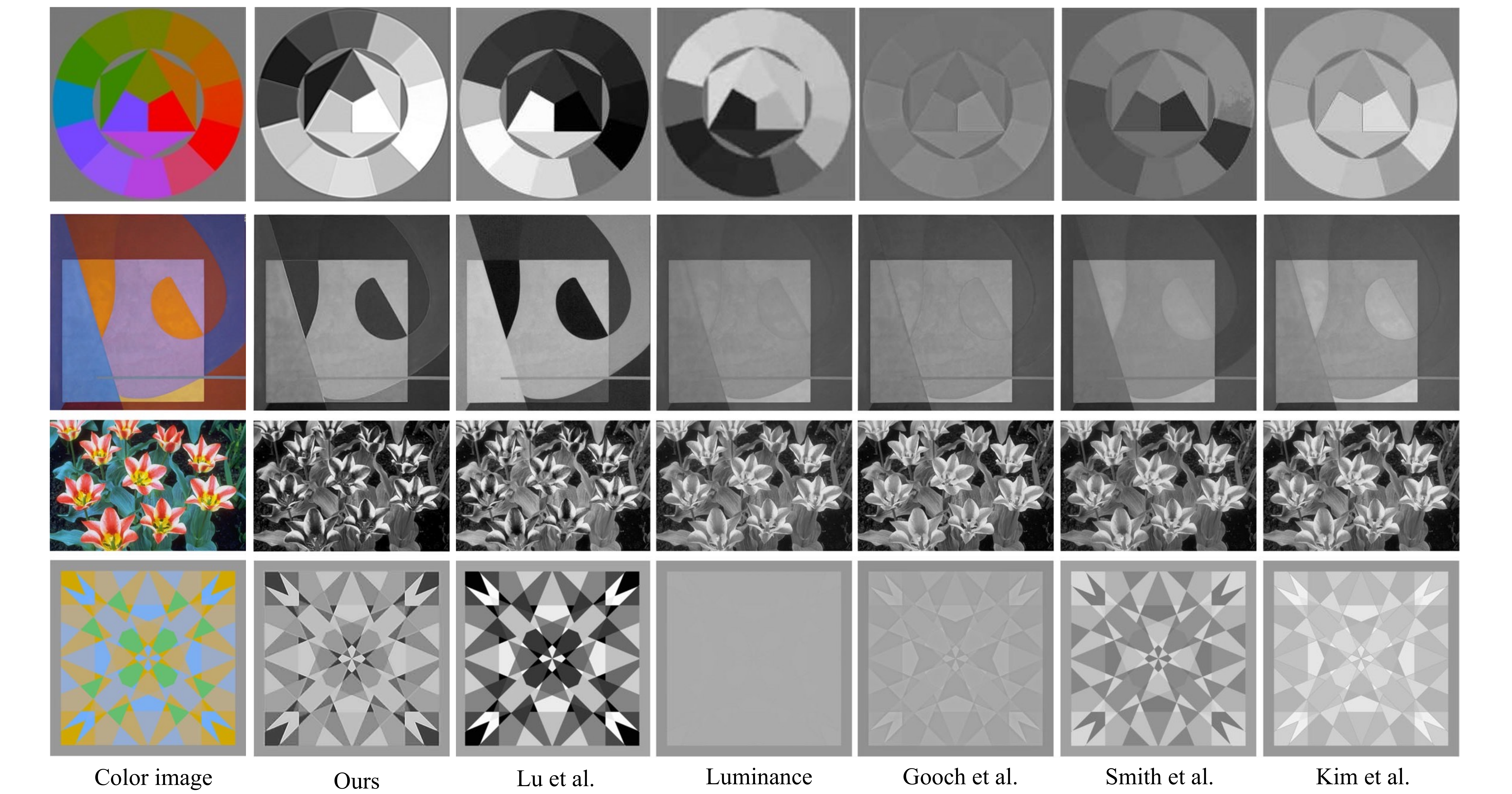}
  \caption{A comparison of decolorized images by different methods. We compare our method trained with conv4\_1 layer with standard luminance and other recent methods \cite{lu2014contrast,gooch2005color2gray,smith2008apparent,kim2009robust}. The results are best viewed electronically.}
  \label{fig:decolorization}
\end{figure*}

\begin{figure*}
\includegraphics[width=\textwidth]{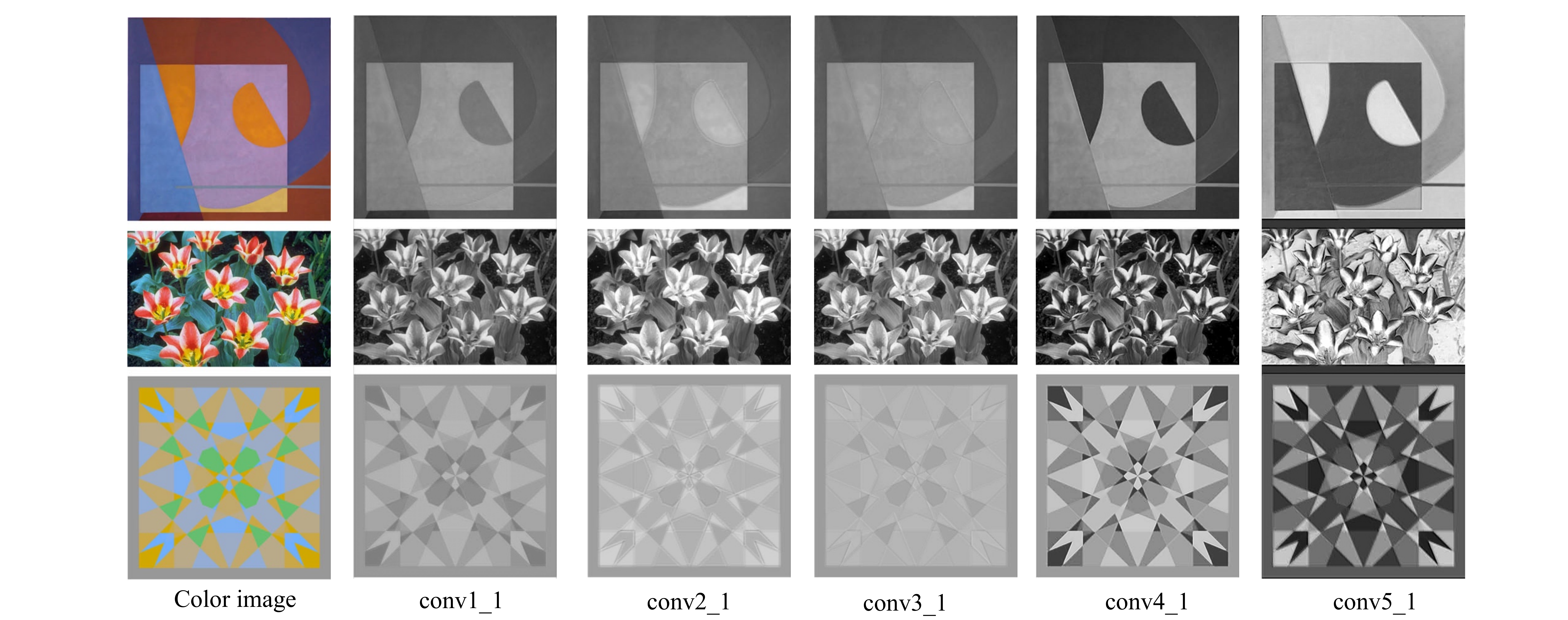}
  \caption{A comparison of decolorization by our methods trained with different level feature perceptual loss. The examples are trained from low level to high level layers in VGGNet. The results are best viewed electronically.}
  \label{fig:decolorization_dfc}
\end{figure*}

\subsection{Image Decolorization.}
Like in image downscaling we also train a two-layer convolutional network to transform color images into grayscale images using the DFC-DIT framework. 
One of the major problems in traditional approach to this task is that in iso-luminant areas the color contrasts will disappear in the grayscale image because even though the pixels have different colors their luminance levels are the same. 
In our neural network based nonlinear mapping framework, we enforce deep feature consistency which means that the spatial correlations of the color images are preserved in the grayscale image. Thus even in iso-luminant regions, the color contrasts will be preserved as grayscale contrasts. 

\textbf{Qualitative results.}
Again, our fully convolutional neural network architecture can be applied to process images of any sizes even though the training images have a fixed size. 
Fig. \ref{fig:decolorization} shows several comparative results against standard luminance and recent color to grayscale methods \cite{lu2014contrast,gooch2005color2gray,smith2008apparent,kim2009robust}. Our training-based approach can preserve the color contrasts of the original images, the grayscale images appear sharp and fine details are well protected. It is interesting to note that unlike previous methods, we did not explicitly compute color contrasts and grayscale contrasts, instead we only enforce deep feature consistency of the color and the decolorized images. From these examples, we have shown convincingly that our DFC-DIT framework is an effective decolorization method.  

\begin{figure*}
\includegraphics[width=\textwidth]{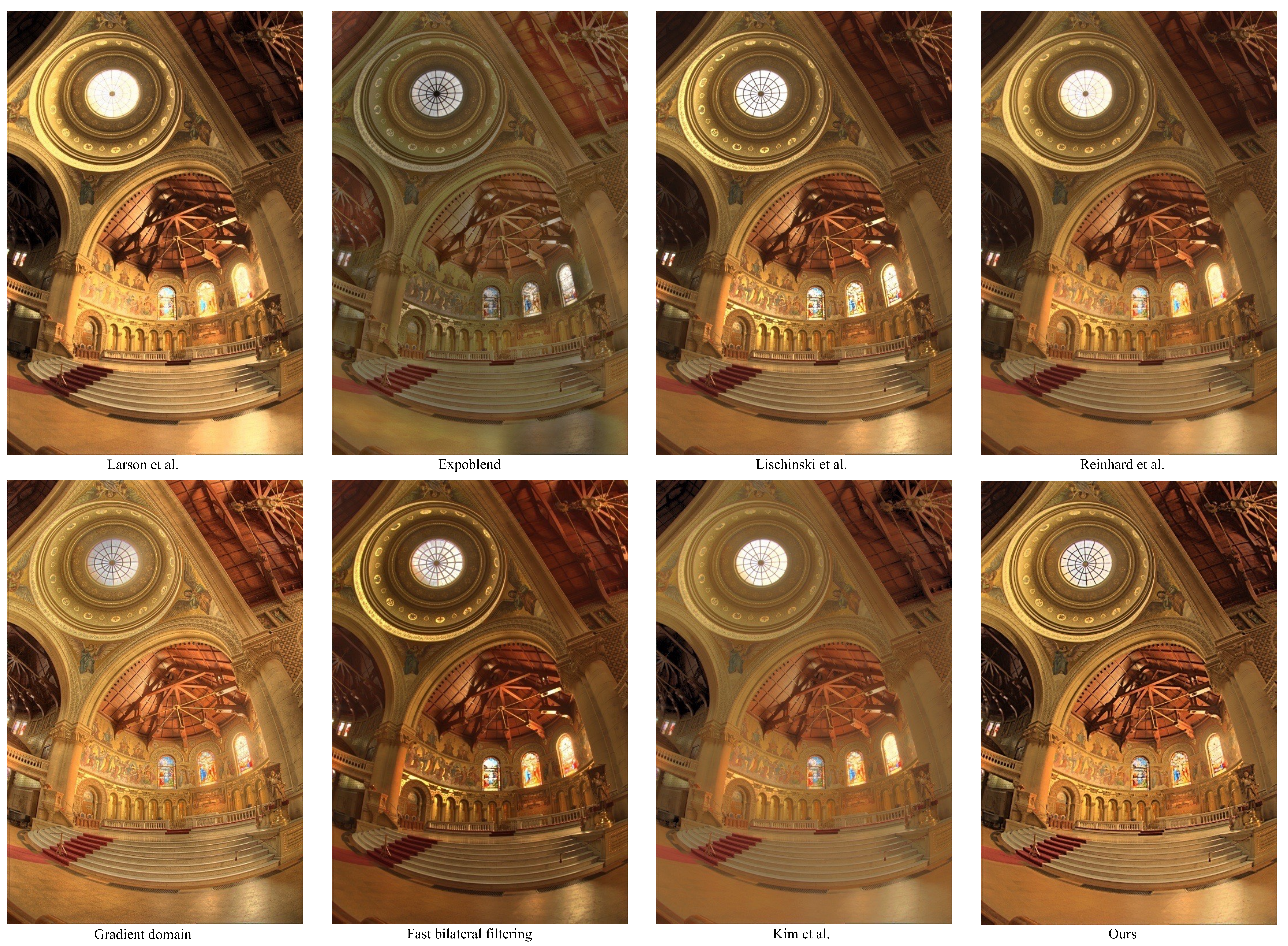}
  \caption{Stanford Memorial Church displayed using different methods. We show those of Larson et al. \cite{larson1997visibility}, Expoblend \cite{bruce2014expoblend}, Lischinski et al. \cite{lischinski2006interactive}, Reinhard et al. \cite{reinhard2002photographic}, gradient domain \cite{fattal2002gradient}, fast bilateral filtering \cite{durand2002fast} and Kim et al. \cite{Kim2008}. Our results are based on feature perceptual loss of 3 layers conv1\_1, conv2\_1 and conv3\_1.}
  \label{fig:hdr}
\end{figure*}

\begin{figure*}
\includegraphics[width=\textwidth]{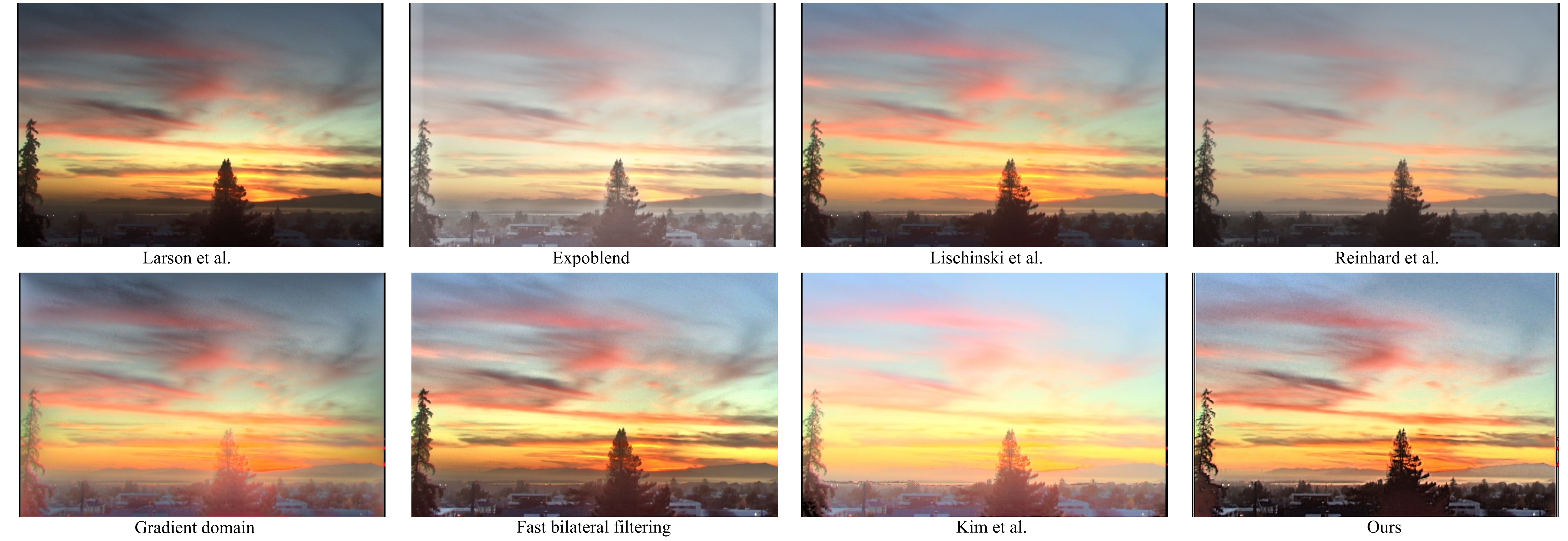}
  \caption{Sunset image displayed using different methods. We show those of \cite{larson1997visibility}, Expoblend \cite{bruce2014expoblend}, Lischinski et al. \cite{lischinski2006interactive}, Reinhard et al. \cite{reinhard2002photographic}, gradient domain \cite{fattal2002gradient}, fast bilateral filtering \cite{durand2002fast} and Kim et al. \cite{Kim2008}. Our results are based on feature perceptual loss of 3 layers conv1\_1, conv2\_1 and conv3\_1. Our results are based on feature perceptual loss of 3 layers conv1\_1, conv2\_1 and conv3\_1.}
  \label{fig:hdr1}
\end{figure*}

\begin{figure*}
\includegraphics[width=\textwidth]{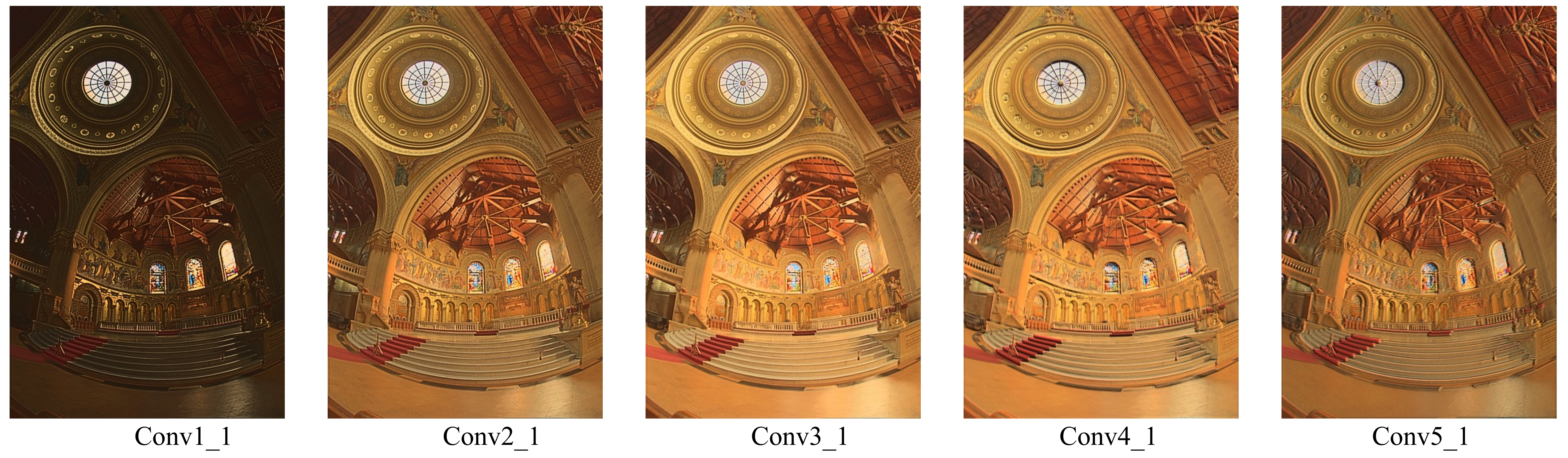}
  \caption{A comparison of HDR image tone mapping by our methods trained with different level feature perceptual loss. The results are best viewed electronically.}
  \label{fig:hdr_dfc}
\end{figure*}

\begin{figure*}
\includegraphics[width=\textwidth]{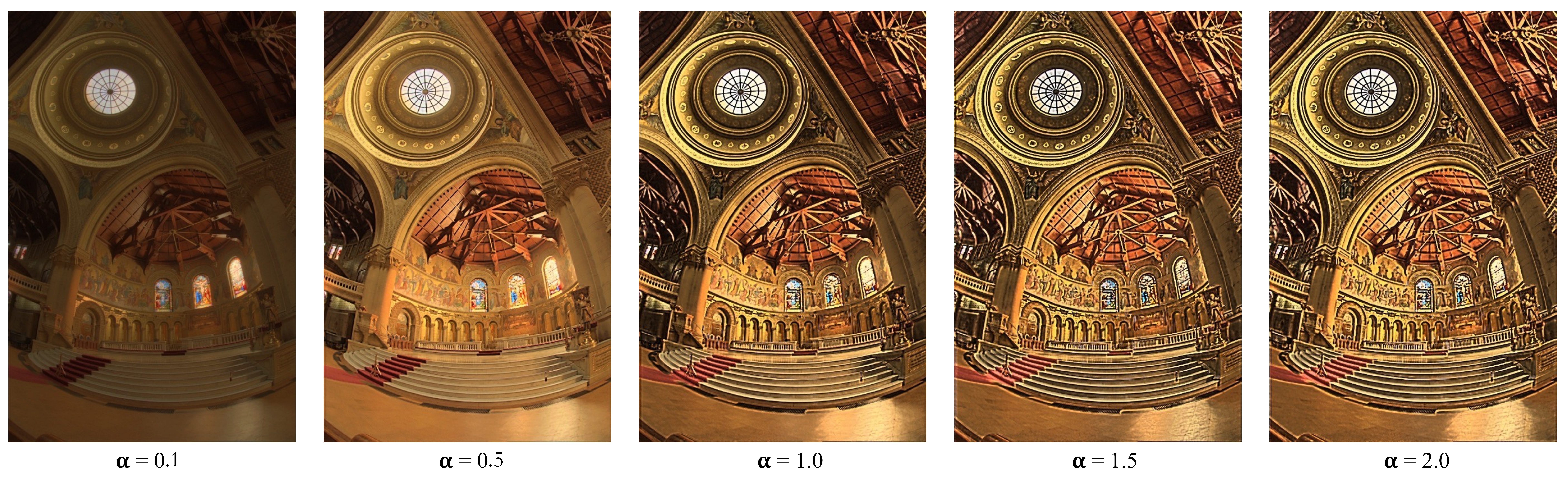}
  \caption{A demonstration of the effects of logarithmic compression based on feature perceptual loss of 3 layers conv1\_1, conv2\_1 and conv3\_1.}
  \label{fig:hdr_logscale}
\end{figure*}

\textbf{Deep feature consistency at different layers.} We also conduct experiments to evaluate how deep feature consistency at different hidden layer of the loss network affects the decolorization results. 
Results produced by models trained with perceptual loss of different hidden layers are shown in Fig. \ref{fig:decolorization_dfc}. Again we can see that all the transformed images are able to reconstruct the content of the original color image and preserve the contrasts. Compared to lower layers, the decolorized images from higher layers do a better job at reconstructing fine details, especially the contrast preservation that is desired. Specifically the results from lowest layers, i.e., conv1\_1 are similar to the luminance channel (Fig. \ref{fig:decolorization}), isoluminant regions are mapped onto the same output intensity and global appearance is not well preserved. Constructing feature perceptual loss from higher layer is better for contrast preserving. However when using the highest conv5\_1 layer (Fig. \ref{fig:decolorization_dfc}), the contrast of the outputs is too high that makes the decolorized images look unnatural. Our best model is trained by using conv4\_1 layer.

\subsection{HDR Image Tone Mapping.}
Unlike in image downscaling and decolorization where a single model is trained offline using a large collection of training images and used to process all testing images, we adapt one network to a single HDR image due to the lack of large HDR dataset available for training. This can be seen as an online process where we use an HDR image's own data to optimize its own transformation function. It is important to note that this approach is realistic in practice as the process only needs the HDR input to produce its tone mapped output and there is no need to use any other extra information. The only slight disadvantage is that it requires online training the neural network using an HDR image's own data before outputting the final tone mapped image. Comparing with training the model offline using a large collection of training images, this online approach will be slower because it needs to adapt the neural network to the current testing image before producing the output tone mapped image. In our implementation on a machine with an Intel Core i7-4790K CPU and a Nvidia Tesla K40 GPU, it takes around 20 seconds to tone map a 768 x 512 HDR image.

It is a common practice to process the HDR radiance map in the logarithmic domain, we feed the logarithm of the radiance signal to the transformation CNN. Dynamic range compression is achieved by a $Tanh$ function in the last layer of the transformation network (Fig. \ref{fig:architecture}(c)). In practice, the dynamic range of the input HDR radiance signal is compressed to the displayable range $[0, 255]$. 
Following the principle of DFC-DIT, the HDR tone mapping transformation network is optimized by enforcing deep feature consistency between the transformation output image and the original HDR radiance map. 

\textbf{Rendering display image.} The output of the transformation network will have the correct dynamic range suitable for display, however, its colour may not be correct due to the nonlinear mapping operations of the transformation CNN. We therefore need to render the output of the transformation network to have the correct colour for display. As in other tone mapping method \cite{qiu2007learning}, the final tone mapped image is rendered as


\begin{align}
\small
R_{out} =  \big(\frac{R_{in}}{L_{in}} \big)^{\gamma} L_{out} \\
G_{out} =  \big(\frac{G_{in}}{L_{in}} \big)^{\gamma} L_{out} \\
B_{out} =  \big(\frac{B_{in}}{L_{in}} \big)^{\gamma} L_{out}
\label{eq:gamma}
\end{align}

where $R_{out}$, $G_{out}$ and $B_{out}$ are the final tone-mapped RGB channels, $R_{in}$, $G_{in}$ and $B_{in}$ are the original radiance values in the corresponding HDR channels, and $\gamma$ can be used to render the correct display colour. $L_{in}$ and $L_{out}$ are respectively the luminance value of the HDR radiance map and the luminance value of the transformation image by the transformation CNN. According to the literature, $\gamma$ should be set between 0.4 and 0.6 and we set it to 0.5 in all our results.


\textbf{Qualitative results.}
Fig. \ref{fig:hdr} shows examples of tone mapping results of some HDR radiance maps of real scenes that are widely used in the literature, i.e., ``Stanford Memorial Church'' and  ``Vine Sunset''. We compare our results with some of the best known and latest methods in the literature including Larson et al. \cite{larson1997visibility}, Expoblend \cite{bruce2014expoblend}, Lischinski et al. \cite{lischinski2006interactive}, Reinhard et al. \cite{reinhard2002photographic}, gradient domain \cite{fattal2002gradient}, fast bilateral filtering \cite{durand2002fast} and Kim et al. \cite{Kim2008}. 
From Fig. \ref{fig:hdr} and Fig. \ref{fig:hdr1}, we can see that our method is able to render the images with excellent visual appearances to keep tiny details and contrast of the radiance map, which are at least as good as those produced by the best methods. 

\textbf{Deep feature consistency at different layers.}
In Fig. \ref{fig:hdr_dfc} we show how feature perceptual loss from different hidden layers affect the tone mapped images of the DFC-DIT framework for HDR tone mapping. 
Overall the tone mapped images based on perceptual losses from the middle level (conv2\_1 and conv3\_1) have a good balance between local and global contrasts. Combining the perceptual losses of first several layers together tend to produce somewhat better results than using a single layer. The tone mapped outputs based on higher layers (conv4\_1 and conv5\_1) appear slightly bumpy effect on different regions.


\textbf{The effects of logarithmic compression.}
As mentioned above, we first compress the HDR radiance map with the logarithmic functions and try to seek the deep feature consistency in the logarithmic domain. We can multiply the compressed radiance map with a factor $\alpha$ to control the logarithmic transformation. The tone mapping results with different $\alpha$ are shown in Fig. \ref{fig:hdr_logscale}. It can be seen that a higher $\alpha$ can lead to a more noticeable local contrast and crisp appearance of tone mapped results. This is because the compressed HDR radiance map with a higher $\alpha$ retains a higher dynamic range in logarithmic domain and retain more important details. It is clear that our method can extract exquisite details from high-contrast images. It works well when $\alpha$ is around 0.5 in our experiments.

\begin{figure*}[htp]
\centering
\includegraphics[width=.31\textwidth]{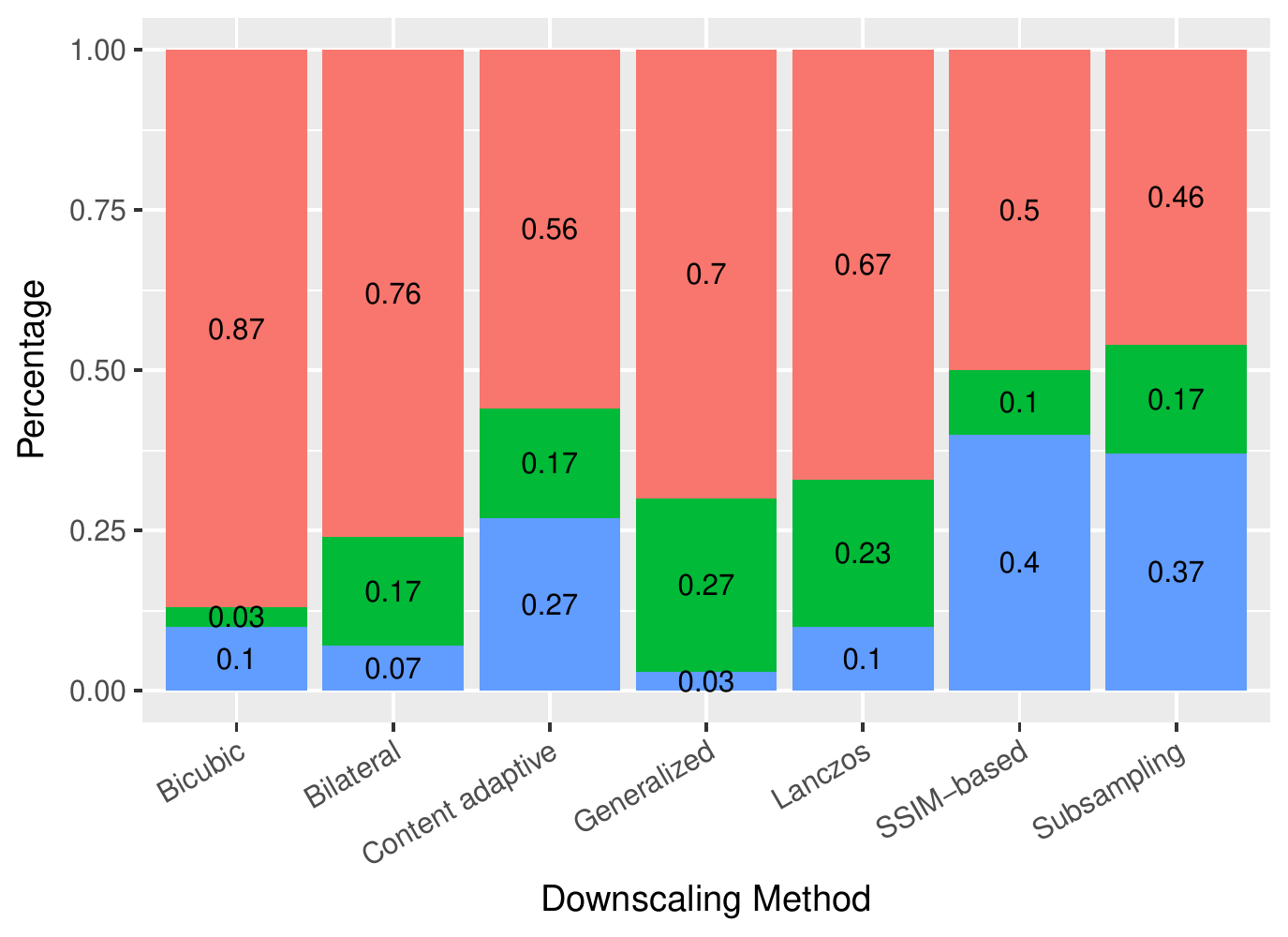}\hfill
\includegraphics[width=.31\textwidth]{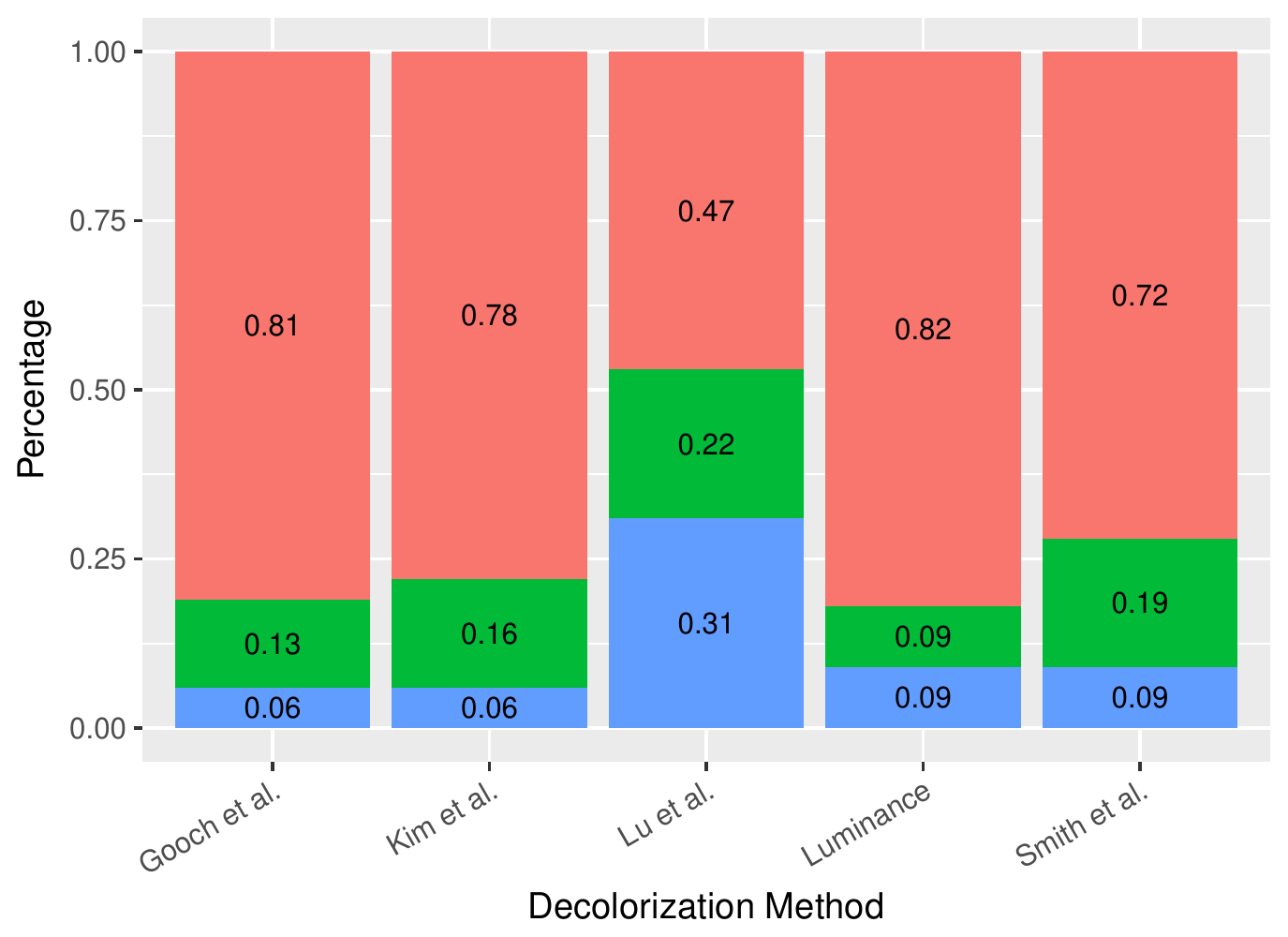}\hfill
\includegraphics[width=.31\textwidth]{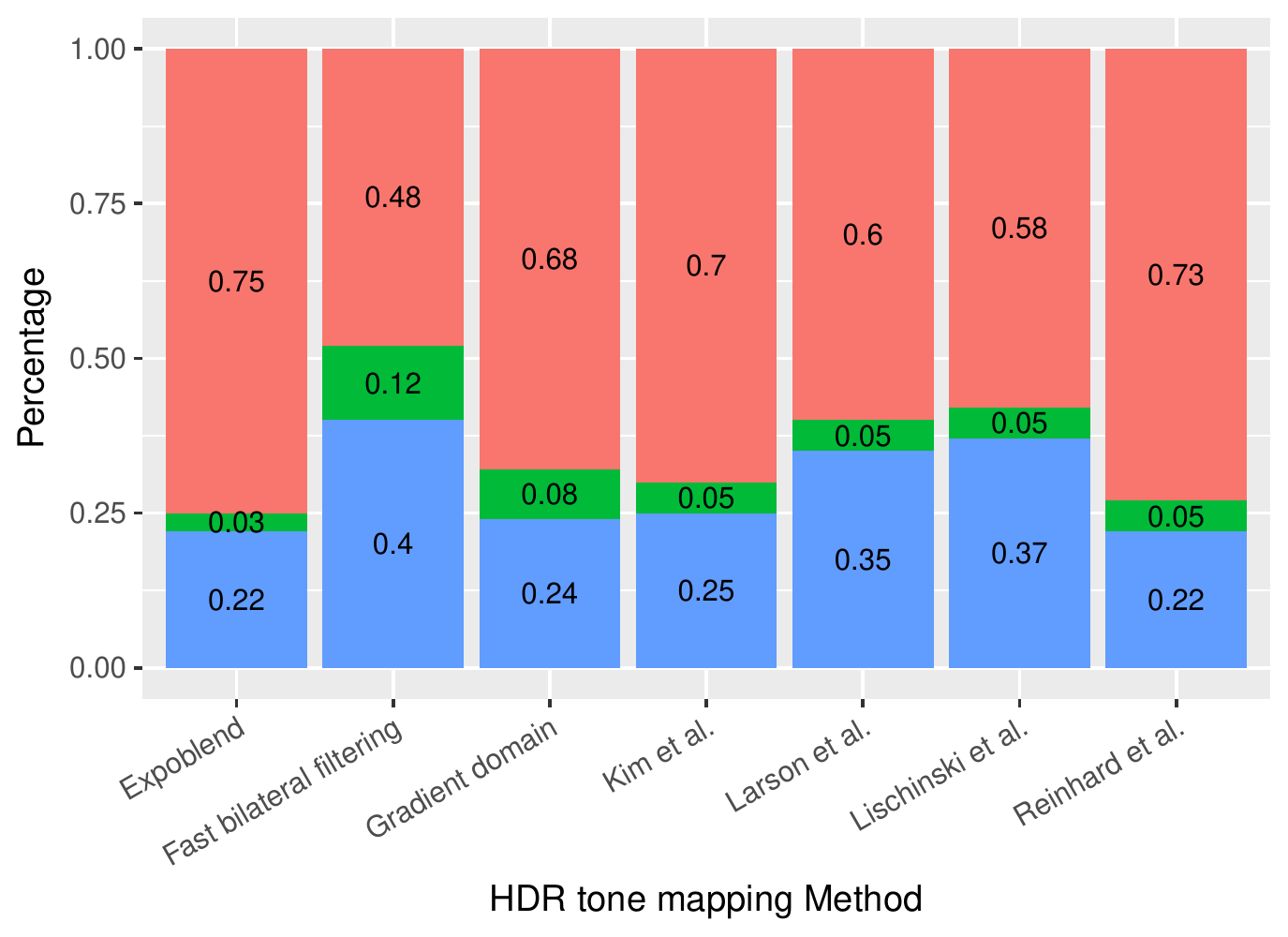}\hfill
\includegraphics[width=.05\textwidth]{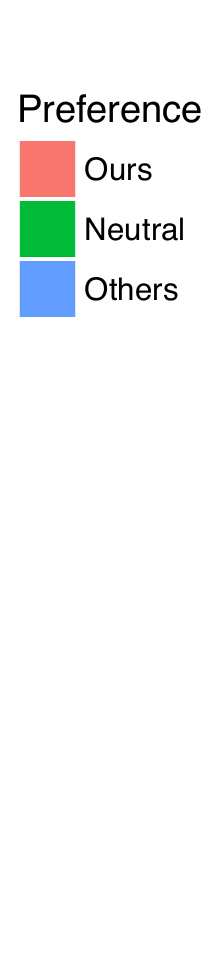}
\caption{Subjective evaluation results. The red areas represent the percentage that our algorithm is selected, green areas for no preference and the blue ones for the other methods.}
\label{fig:subjective_evaluate}
\end{figure*}

\subsection{Subjective Evaluation of DFC-DIT Framework.}

We have conducted a subjective evaluation of results of downscaling, decolorization and HDR tone mapping of the new DFC-DIT framework. For each transformation, we evaluate our technique against several best techniques in the literature. For downscaling, we use bicubic, bilateral, lanczos, subsampling, generalized sampling \cite{nehab2011generalized}, content-adaptive \cite{kopf2013content} and SSIM based method \cite{oeztireli2015perceptually} as the benchmarks. For decolorization, we use luminance the methods of Smith et al. \cite{smith2008apparent}, Kim et al. \cite{kim2009robust}, Gooch et al \cite{gooch2005color2gray} and Lu et al. \cite{lu2014contrast} as benchmarks. For HDR tone mapping we use Larson et al. \cite{larson1997visibility}, fast bilateral filtering \cite{durand2002fast}, gradient domain \cite{fattal2002gradient}, Expoblend \cite{bruce2014expoblend}, Kim et al. \cite{Kim2008}, Lischinski et al. \cite{lischinski2006interactive} and Reinhard et al. \cite{reinhard2002photographic} as benchmarks.
For each image, we show the original input image (in the case of HDR tone mapping, the original radiance map cannot be shown), a version produced by our method and a version of one benchmark technique to subjects and ask which version they prefer or indicate no preference. 50 undergraduate science and engineering students from our university evaluated 10 pairs of images for image downscaling and 8 pairs of images for image decolorization and HDR tone mapping. Fig. \ref{fig:subjective_evaluate} shows the voting results. We can see that there is an obvious preference for our method against all other methods for all the transformation tasks. These results demonstrate DFC-DIT framework is comparable to or better than state-of-the-art techniques. In image downscaling, subsampling and SSIM-based are two competing methods to produce sharp and crisp downscaled images, however subsampling sometimes suffer strong aliasing artifacts like the hair in Fig. \ref{fig:downscaling_other}. In image decolorization, the method of Lu et al. \cite{lu2014contrast} is the best competing candidate that maximally preserves color contrast. However some participants prefer ours than theirs because the decolorized versions of Lu et al. \cite{lu2014contrast} may show too strong contrast while the corresponding color images in fact have low contrasts. For HDR image tone mapping, fast bilateral filtering \cite{durand2002fast} is the best comparable tone mapping operator in our study.

\section{Concluding remarks}

This paper has successfully introduced the DFC-DIT framework which unifies several common difficult image processing tasks. This is also the first time that deep learning has been successfully applied to image downscaling, decolorization and high dynamic range image tone mapping. Experimental results have demonstrated the effectiveness of the method and its state-of-the-art performances.

One fundamental problem for traditional image transformation tasks like image downscaling, image decolorization and HDR image tone mapping is that the problems are inherently ill-posed, because there is no unique correct ground truth, i.e., they are one-to-many mapping problems. For image downscaling fine details should be preserved from visually ambiguous high-resolution inputs; for image decolorization the gray image should be semantically similar to the original color version and preserve the contrast as much as possible in spite of drastic loss of color information; for HDR image tone mapping we want to compress the scene radiance to displayable range while preserving details and color appearance to appreciate the original scene content. Therefore, success in these image transformation tasks requires semantic and perceptual reasoning about the input.

It is very difficult to design a numerical image quality metric to measure the perceptual similarity between the transformed outputs and the original input. Based on two crucial insights into the determining factors of visual quality of images and the properties of deep convolutional neural network, we have developed the deep feature consistent deep image transformation (DFC-DIT) framework which unifies common and ill-posed image processing tasks like downscaling, decolorization and HDR tone mapping. We have shown that the hidden layer outputs of a pretrained deep CNN can be used to compare perceptual similarities between the input and the output images. One possible explanation is that the hidden representations of a pre-trained CNN have captured essential visual quality details such as spatial correlation information and other higher level semantic information. Exactly which hidden layer represents what kind of essential visual quality details is not very clear and we have shown that perceptual losses constructed with different hidden layer features can affect the final results. Future researches are needed to understand better the kinds of visual semantics captured by the hidden features of the pre-trained CNN in the context of the DFC-DIT framework. A better way to combine (e.g. weighting) different level deep features may lead to better and more consistent results.

\ifCLASSOPTIONcaptionsoff
  \newpage
\fi



%

\bibliographystyle{IEEEtran}
\bibliography{IEEEabrv,IEEEexample}

%








\end{document}